\documentclass[nohyperref]{article}

\usepackage{microtype}
\usepackage{graphicx}
\usepackage{subfigure}
\usepackage{booktabs} 
\usepackage{paralist}

\usepackage{hyperref}

\usepackage[accepted]{icml2022}

\usepackage{amsmath}
\usepackage{amssymb}
\usepackage{mathtools}
\usepackage{amsthm}

\usepackage[capitalize,noabbrev]{cleveref}

\theoremstyle{plain}
\newtheorem{theorem}{Theorem}[section]

\theoremstyle{definition}
\newtheorem{definition}[theorem]{Definition}

\theoremstyle{remark}

\usepackage[textsize=tiny]{todonotes}

\icmltitlerunning{Collective Obfuscation and Crowdsourcing}

\begin{document}

\twocolumn[
\icmltitle{Collective Obfuscation and Crowdsourcing}




\begin{icmlauthorlist}
\icmlauthor{Benjamin Laufer}{cornell}
\icmlauthor{Niko A. Grupen}{cornell}
\end{icmlauthorlist}

\icmlaffiliation{cornell}{College of Computing and Information Sciences, Cornell University}

\icmlcorrespondingauthor{Benjamin Laufer}{bdl56@cornell.edu}
\icmlcorrespondingauthor{Niko Grupen}{nag83@cornell.edu}

\icmlkeywords{Machine Learning, Obfuscation, Crowdsourcing, Reporting, Platform, Misinformation, Spam}

\vskip 0.3in
]



\printAffiliationsAndNotice{\icmlEqualContribution} 

\begin{abstract}
Crowdsourcing technologies rely on groups of people to input information that may be critical for decision-making. This work examines \textit{obfuscation} in the context of reporting technologies. We show that widespread use of reporting platforms comes with unique security and privacy implications, and introduce a threat model and corresponding taxonomy to outline some of the many attack vectors in this space. We then perform an empirical analysis of a dataset of call logs from a controversial, real-world reporting hotline and identify coordinated obfuscation strategies that are intended to hinder the platform’s legitimacy. We propose a variety of statistical measures to quantify the strength of this obfuscation strategy with respect to the structural and semantic characteristics of the reporting attacks in our dataset.
\end{abstract}

\section{Introduction}
Crowdsourcing is a method of information retrieval that relies on the public to supply information to decision-makers including state authorities. In the public reporting domain, even established systems, such as emergency reporting (e.g. 911) and information hotlines (e.g. 311) are not exempt from problems of distrust \cite{rock2019one, sasson2015barriers, kessell2009effect, smith2003community, clark2020advanced}. It is no surprise then that more controversial platforms---including Victims of Immigration Crime Engagement (VOICE) \cite{kopan2017voice} and Texas Right to Life \cite{mccammon2021texas}---are subjected to unintended use, spamming attacks, false reports, DDoS attacks, and more, which collectively render their crowdsourced information useless. These attacks, however, differ in nature from standard security and privacy attacks on technology platforms, as they represent a form of \textit{obfuscation}; defined as ``the deliberate addition of ambiguous, confusing, or misleading information to interfere with surveillance and data collection" \cite{brunton2013political, brunton2015obfuscation}.

We posit that, unlike studies that characterize security with respect to the \textit{intended use} of a technology, the reporting domain requires consideration of both \textit{harmful use} and \textit{legitimate misuse} (e.g. obfuscation). To this end, we introduce a parallel threat model describing both threats to the platform's legitimate functioning and platform-enabled violence. Inspired by \citet{thomas2021sok}, we propose a taxonomy of use and abuse in how people interact with reporting platforms, and discuss how legitimate responses to reporting platforms depend significantly on social context. 

Finally, we study obfuscation in the context of a real-world collective spamming attack on the VOICE reporting system. Using open-source data from the VOICE system's call logs, we analyze both structural (e.g. report length) and semantic (e.g. sentence embedding distances) properties of spam and non-spam reports. We propose a variety of statistical measures to quantify the strength of an obfuscation strategy with respect to similarity and dissimilarity in the high-dimensional embedding spaces generated by neural network language models. Our analysis reveals a number of interesting insights for spamming techniques on reporting technologies: 
\begin{inparaenum}[(i)]
    \item Spam reports are longer than true reports (intending to waste operator time);
    \item Spam reports are semantically \textit{disparate}, covering a wider swath of topics and clustering into more disjoint groups;
    \item Spam reports are thematically similar, suggesting that they may be from coordinated Internet campaigns.
\end{inparaenum}
While many of the spam messages in the VOICE dataset were easily identifiable, future deception campaigns using advanced language models may pose more significant threats because they will more closely resemble the distribution of true reports.

\section{Taxonomy of Threats}
\begin{figure}[t!]
    \centering
    \vspace{-.05in}
    \makebox[\linewidth][c]{\includegraphics[width=\linewidth]{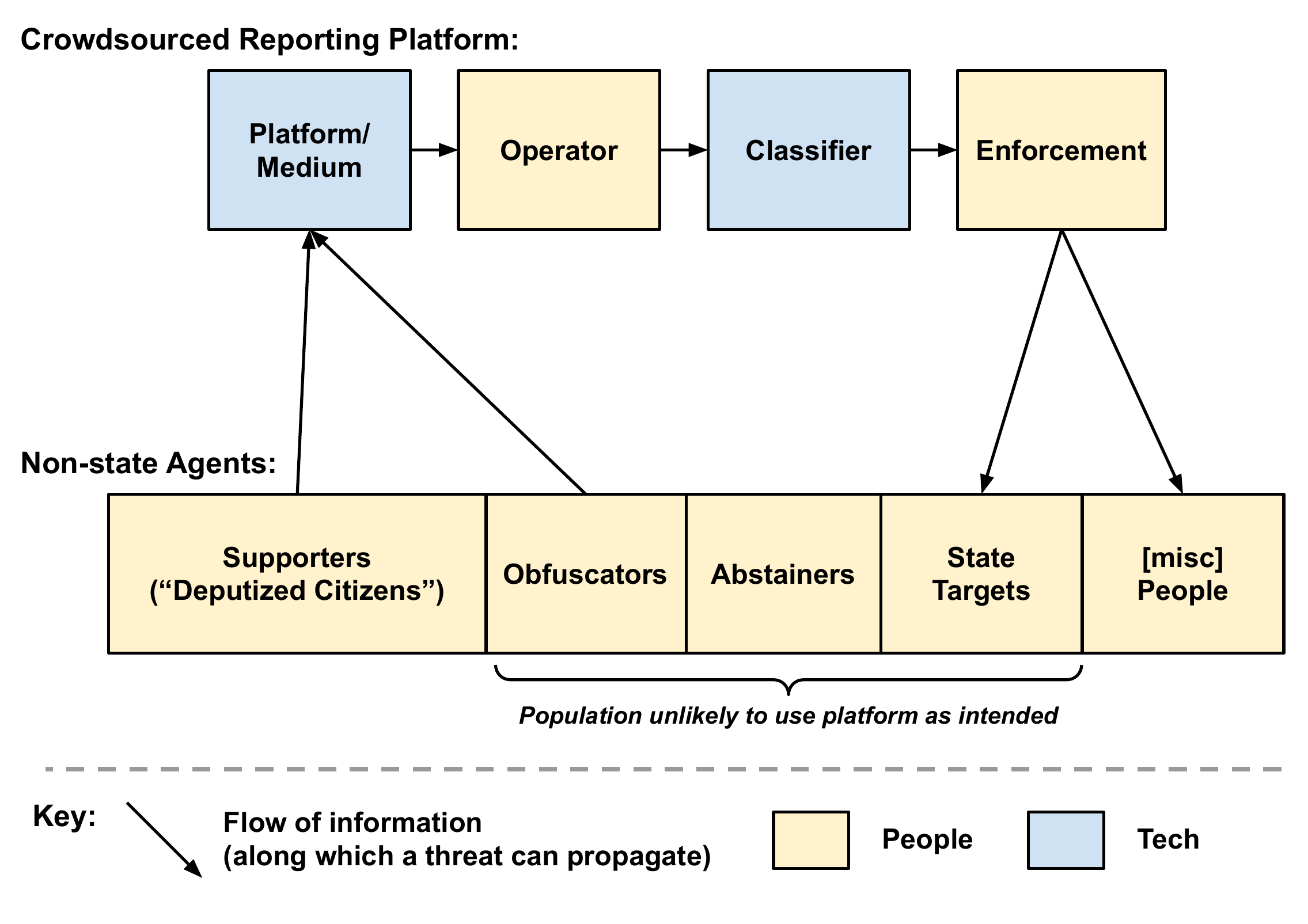}}
    \vspace{-.2in}
    \caption{Diagram representing the flow of information through reporting platforms.}
    \label{platformflowchart}
    \vspace{-.1in}
\end{figure}

In this section, we report a taxonomy of threats that are mediated or enabled by reporting platforms. These threats are accumulated through systematically mapping the information flows that constitute reporting platforms (Figure \ref{platformflowchart}). 

\vspace{-.2in}
\subsection{Deceptive Reporting}
Submitting a deceptive report on a reporting platform entails fooling a call operator or spam classifier in order to shift the functionality and resource allocation associated with the platform. These reports may erode the reporting platform’s legitimacy and functionality. Deceptive reports fall into two broad categories: false reporting aimed at harming or disempowering another private individual (described in \citet{thomas2021sok}) and false reporting aimed at harming or disempowering the reporting platform itself. 

\subsubsection{Weaponizing Reports} Deceptive reporting to intimidate, disempower or harm typically involves a private-citizen attacker whose goal is to harm a private-citizen target. In a phenomenon known as “SWATing”, individuals sometimes use false reports to provoke emergency services to locate and confront a target. If a caller falsely reports a serious crime (for exmaple a bomb threat), use-of-force may lead to fatality \cite{krebs2019swat}. Attackers who weaponize reports have goals ranging from intimidation to coercion to direct physical harm. Depending on the purpose and functionality of a reporting system, weaponized reports can intimidate targets or convey a threat.

\vspace{-.1in}
\subsubsection{Poison Reports} Poison reports aim to harm platforms rather than private individuals. Analogous to cache poisoning \cite{son2010hitchhiker}, poison reports are a particular type of spam that relies on deceiving an operator or spam classifier. By planting false reports that are believed to be true by the platform and its operators, poison data may significantly harm the platform’s ability to distinguish true reports \cite{vincent2021data}. These attacks are particularly harmful in large (variegated) quantities, or when a false positive report classification incurs large costs for the platform operator.
 
\subsubsection{Exaggerated Reports} These reports come from people who intend to use the reporting platform for its express purposes, but wish to receive priority. Participants may behave \textit{strategically}, knowing that their case may be prioritized if they exaggerate the urgency of their report. This can take the form of an emergency declaration, use of inflammatory or urgent words, pleading, or fabricating aspects of an account. 
 
\subsection{Abnormal Reporting}
Abnormal reports contain information that was not expected or designed to be processed by the platform. These reports need not be deceptive or even malicious, but can be considered ``threats” because they harm the functioning of a reporting system. The mechanisms through which they cause harm include creating many meaningless reports that need to be sifted through, wasting phone operators’ time, creating backlogs, and generally adding friction to the reporting process. Abnormal reports span a wide number of different categories; we list a few below.

\subsubsection{Opinions} Crowdsourced reporting platforms, especially non-emergency platforms, can be inundated with calls from people who simply wish to state sentiments and opinions. For example, people may call to state their approval or disapproval of a platform’s existence. In this case, the ‘attacker’ is an individual user, and the target is the reporting platform and its operators who read or listen to reports.

\subsubsection{Trolling} We define trolling as the use of ``inflammatory, insincere, digressive, extraneous, or off-topic messages” \cite{wikitroll} to provoke or manipulate. Troll behavior is distinct from deceptive reporting---although troll reports are often false, they are not deceptive nor do they expect to successfully trick an operator or platform into reallocating resources. Attackers target an operator or platform and might be capable of coordinating large-scale attacks on platforms \cite{birkbak2018, navarro2021trolls}.

\subsubsection{Threats} These reports are often politically-motivated reports by individuals who believe the platform is harmful or should not exist. Their goal is to intimidate, coerce, or threaten the humans who are operating and maintaining the reporting system. Attacks range from accusatory political sentiments (e.g., telling an operator that they should be ashamed of their work) to direct threats (e.g., attempting to use an operator's personal information to intimidate).

\subsubsection{Hate Speech and Profanity} The use of corrosive, belligerent, and hateful language can occur on any platform that is not censored. These reports need not be relevant to the platform’s intended purpose, and hateful language may not be directed at anyone in particular. There may be no real target or goal (from a security perspective) associated with these attacks, but they may harm a platform’s functioning and so should be taken seriously.

\subsubsection{Accidental Calls} Many people call a hotline by accident. Emergency hotlines are inundated with accidental calls, such as `butt dials' or mistakenly entered area codes or phone numbers \cite{friedman2017punishing}. These calls are not malicious in nature but need to be efficiently triaged to prioritize intended calls.

\subsection{Overloading}
Overloading attacks take aim at a system’s ability to field large numbers of reports or traffic. Overloading attacks specific to reporting platforms include report spamming, raiding or brigading, and distributed denial of service (DDoS). For a general overview of overloading see \citet{thomas2021sok}.

\subsubsection{Report Spam} In this context, we define spamming as friction-generating calls whose purpose is solely to use platform resources. Report spam can include troll reports, but may even include silent calls, music, pre-recorded messages, or operator-operator routing calls (where two operators are put in touch by an intermediary user to sow confusion).

\subsubsection{Raiding or Brigading} These are instances where people coordinate to overwhelm a feed, platform, or comment section to target an individual or group. In reporting software, raiding and brigading may take the form of script-based reports, large-scale spam `dumps', or coordinated trolling.

\subsubsection{Distributed Denial of Service (DDoS)} DDoS attacks make a platform useless by jamming communication channels so that nobody can participate in information reporting. 

\subsection{Information Leakage}
Crowdsourced enforcement mechanisms can enable interpersonal harm, even when they are intended by the system's designer.  In a number of cases including immigration enforcement, policing, hotlines enable people to report one another and leak private information that could be used to incarcerate, interrogate, and search. Reporting platforms can be a means by which people leak citizenship documentation status, criminal histories, or private sexual and health information---all instances of information leakages.

\subsection{Coercion} 
Depending on the decision motivating a crowdsourced information retrieval system, the platforms may enable coercive leverage between people. 911 systems enable people to call the police on anybody on the street or in their lives. This ability injects state force into interpersonal relationships, which can lead to threats and coercion between people. Somebody may threaten to call the police on somebody else, and this type of threat can be harmful and manipulative.

\subsection{Surveillance}
Reporting systems turn everyday citizens into deputies whose responsibility is to report information that may be relevant to state law enforcement, regulatory services, or resource allocation. These systems raise two potential threats in the category of surveillance, which we refer to as \textit{state surveillance} 
and \textit{interpersonal surveillance}. 

\subsubsection{State Surveillance} Reporting platforms can enable state surveillance by expanding the information available to law enforcement. This can provide warrant for searching, spying, and arresting people. Further, anonymity guarantees in a variety of platforms can be violated if law enforcement has a reasonable cause.  

\subsubsection{Interpersonal Surveillance} Reporting platforms rely on people to observe and provide information to platform operators. In the case of law enforcement, citizens feel they have a responsibility to inform authorities of emergency situations. Citizens may take this responsibility too far, and feel emboldened to surveil neighbors and community members who they deem suspect. The same is true for immigration hotlines, abortion hotlines, and even whistleblower and other crowdsourced information hotlines. Such snooping behavior can be a threat to privacy. 

\section{Case Study: VOICE Logs}
In this section, we analyze a dataset of 5164 publicly-available call logs from the VOICE reporting system in 2017, which exhibits some phenomena described above. We focus on the following questions:
\begin{inparaenum}[(i)]
    \item What structural qualities (e.g. report length) differentiate spam from non-spam interactions?
    \item Do spam reports cover a wider range of topics than non-spam reports?
    \item Are spam reports semantically similar to other spam reports?
\end{inparaenum}
We address each of these questions in the subsections below. The structural analysis can be found in Appendix \ref{apdx_struct}.

\subsection{Semantic Analysis}
To capture the semantic content of the reports, we generate sentence-level embeddings for each report using Sentence-BERT \cite{reimers2019sentence}. Sentence embeddings map text sentences into a high-dimensional vector space in which semantically similar sentences are close, and unrelated sentences are far apart. Such representations are effective for downstream textual tasks like clustering and semantic search \cite{conneau2018senteval}. Here, we leverage this high-dimensional semantic space to examine the similarities and dissimilarities of spam vs. non-spam reports by clustering their respective embeddings and measuring text similarity with a variety of metrics that we introduce to examine the underlying report distributions.

\subsubsection{Clustering Analysis}
\begin{figure}[t!]
    \centering
    \makebox[\linewidth][c]{\includegraphics[width=0.95\linewidth]{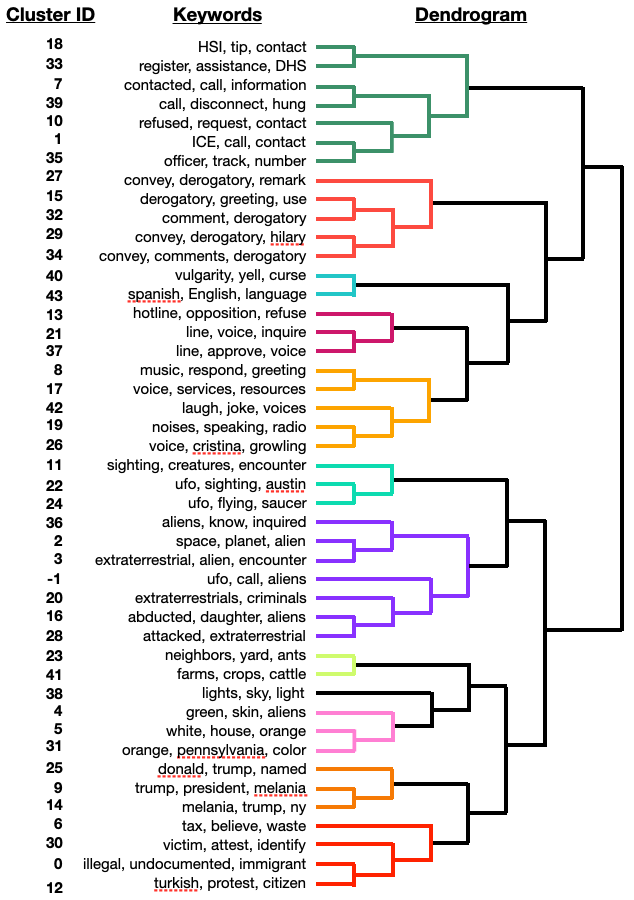}}
    \vspace{-0.075in}
    \caption{Hierarchical clustering of VOICE logs. The first seven clusters (top, colored green) appear to be true reports, whereas many of the others fall into threat categories including opinions (e.g. cluster 13) and obvious spam (e.g. references to UFO's).}
    \label{fig_clustering_all}
    \vspace{-0.075in}
\end{figure}

First, we perform hierarchical clustering over the set of report embeddings. Formally, given a distribution of report sentences $S$, we pass each sentence $s_i \in S$ through an embedding-generating function $f_\theta : S \rightarrow \mathbb{R}^d$ (with parameters $\theta$), which produces a $d$-dimensional embedding vector $v_i \in R^d$. Following the method introduced by \citet{grootendorst2020bertopic}, we cluster the embeddings by preojecting each $v_i$ onto a lower-dimensional manifold using UMAP reduction \cite{mcinnes2018umap}, then performing density-based clustering with HDBSCAN \cite{campello2013density}. The optimal clustering yields a set of topics $T$ for the reports, which are shown in Figure \ref{fig_clustering_all}. From the y-axis of Figure \ref{fig_clustering_all}, we see a number of obviously spam clusters that correspond to attack vectors from our taxonomy; such as abnormal reporting---e.g. false reports describing UFOs and extraterrestrials (trolling), politically-charged commentary (opinions), and operator harassment (threats, profanity)---and overloading---e.g. staying silent or playing music (report spam). This analysis suggests that a large number of spam calls can be accurately identified by topic alone. We therefore conjecture that a cursory human review of the clusters produced yields spam/non-spam labels that are highly accurate at detecting semantically obvious, non-deceptive spam attacks. We supplement this analysis by clustering spam and non-spam reports separately in Appendix \ref{apdx_clustering}.
\label{sec_clustering}

\subsubsection{Semantic Similarity Analysis}
We also quantify the similarity of spam vs. non-spam reports as a function of distance in semantic space. Given spam embeddings $\mathbf{v^{\textrm{\textbf{spam}}}} = \{v_1^{\textrm{spam}}, v_2^{\textrm{spam}}, ..., v_n^{\textrm{spam}}\}$, we define the following three distance measures:
\begin{definition}
Within-category distance $\bar{D}_{\textrm{WC}}$ is the average cosine distance between each spam embedding $v_i^{\textrm{spam}}$ and the mean spam embedding $\bar{v}^{\textrm{spam}}$:
\begin{equation*}
    \bar{D}_{\textrm{WC}}(\mathbf{v^{\textrm{\textbf{spam}}}}) = \frac{1}{n}\sum_i^n 1 - \frac{v_i^{\textrm{spam}} \cdot \bar{v}^{\textrm{spam}}}{||v_i^{\textrm{spam}}||_2 ||\bar{v}^{\textrm{spam}}||_2}
\end{equation*}
\vspace{-0.075in}
\end{definition}
\vspace{-0.075in}

\noindent Intuitively, $\bar{D}_{\textrm{WC}}(\mathbf{v^{\textrm{\textbf{spam}}}})$ measures the extent to which spam reports \textit{resemble each other}. Due to the varied topics of spam, as outlined in Section \ref{sec_clustering}, we expect there to be a greater difference between the semantic content of spam than there is for non-spam. We therefore hypothesize that $\bar{D}_{\textrm{WC}}(\mathbf{v^{\textrm{\textbf{spam}}}}) > \bar{D}_{\textrm{WC}}(\mathbf{v^{\textrm{\textbf{non-spam}}}})$, where $\bar{D}_{\textrm{WC}}(\mathbf{v^{\textrm{\textbf{non-spam}}}})$ is the average distance for non-spam embeddings $\mathbf{v^{\textrm{\textbf{non-spam}}}}$. 
\begin{figure*}[t!]
    \centering
    \includegraphics[width=0.95\textwidth]{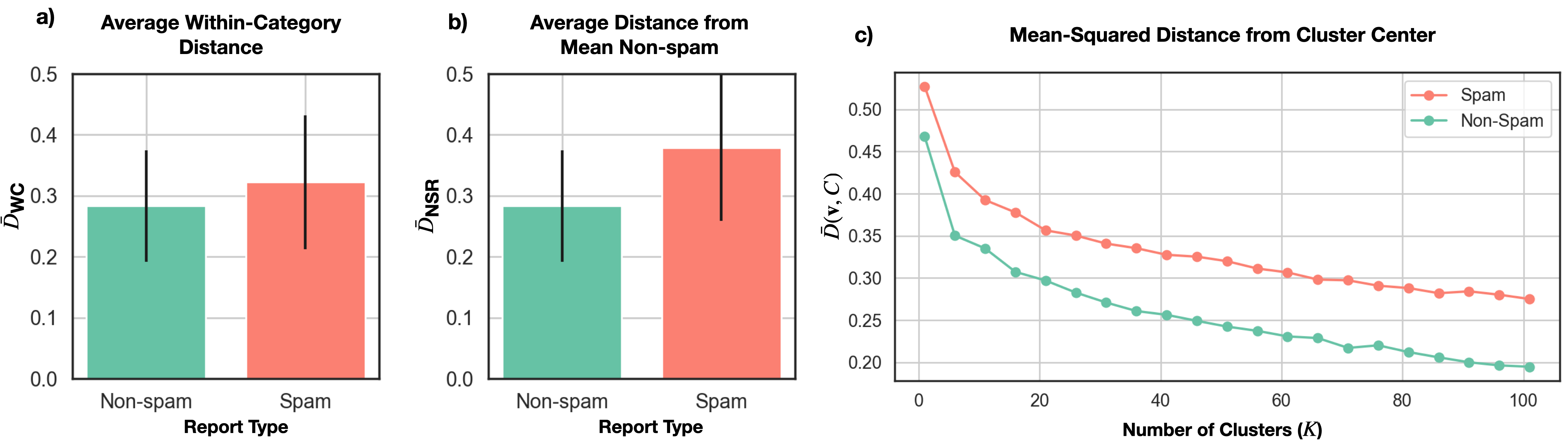}
    \caption{\textbf{a)} Average within-category distance. Spam (red) reports display less semantic similarity than non-spam (green) reports, but not considerably less. Despite covering a wide range of topics, spammers are relatively within each topic. \textbf{b)} Average distance from the mean non-spam embedding. Spam are further from the mean non-spam vector than non-spam, indicating that spam reports are not well-disguised. \textbf{c}) Distance of spam vs. non-spam samples from their cluster centers as a function of the number of clusters $K$.}
    \label{fig_distances}
\end{figure*}
\newline 

\begin{definition}
Distance from the mean non-spam report $\bar{D}_{\textrm{NSR}}$ is the average cosine distance between each spam embedding $v_i^{\textrm{spam}}$ and the mean non-spam embedding $\bar{v}^{\textrm{non-spam}}$:
\begin{equation*}
    \bar{D}_{\textrm{NSR}}(\mathbf{v^{\textrm{\textbf{spam}}}}) = \frac{1}{n}\sum_i^n 1 - \frac{v_i^{\textrm{spam}} \cdot \bar{v}^{\textrm{non-spam}}}{||v_i^{\textrm{spam}}||_2 ||\bar{v}^{\textrm{non-spam}}||_2}
    \vspace{-0.025in}
\end{equation*}
\end{definition}
\vspace{-0.025in}
\noindent Intuitively, $\bar{D}_{\textrm{NSR}}(\mathbf{v^{\textrm{\textbf{spam}}}})$ measures the extent to which spam reports \textit{resemble non-spam reports}. If $\bar{D}_{\textrm{NSR}}(\mathbf{v^{\textrm{\textbf{spam}}}}) \approx \bar{D}_{\textrm{WC}}(\mathbf{v^{\textrm{\textbf{non-spam}}}})$, we can say that the generated spam reports are well-disguised semantically amongst non-spam reports. For spam VOICE reports, which cover a much wider qualitative range of semantic content than non-spam reports in Figure \ref{fig_clustering_all}, we expect $\bar{D}_{\textrm{NSR}}(\mathbf{v^{\textrm{\textbf{spam}}}}) > \bar{D}_{\textrm{WC}}(\mathbf{v^{\textrm{\textbf{non-spam}}}})$.
\newline 

\begin{definition}
Let $C^{\textrm{spam}}$ be the set of $K$ disjoint clusters (with centroids $\mu_j^{\textrm{spam}}$) computed by the K-means algorithm \cite{hartigan1979algorithm} over embeddings $\mathbf{v^{\textrm{\textbf{spam}}}}$. We define the mean-squared distance of each sample from its cluster centroid, $\bar{D}(\mathbf{v^{\textrm{\textbf{spam}}}}, \mu_j^{\textrm{spam}})$, as:
\begin{equation*}
    \bar{D}(\mathbf{v^{\textrm{\textbf{spam}}}}, \mu_j^{\textrm{spam}}) = \frac{1}{n}\sum_i^n ||v_i^{\textrm{spam}} - \mu_j^{\textrm{spam}}||^2
    \vspace{-0.075in}
\end{equation*}
\end{definition}

\noindent Intuitively, $\bar{D}(\mathbf{v^{\textrm{\textbf{spam}}}}, \mu_j^{\textrm{spam}})$ measures the \textit{internal coherence} of the clusters generated for spam reports. Given the large variety of spam observed, we expect that it will be much more difficult to cluster coherently than non-spam, so we hypothesize that $\bar{D}(\mathbf{v^{\textrm{\textbf{spam}}}}, \mu_j^{\textrm{spam}}) > \bar{D}(\mathbf{v^{\textrm{\textbf{non-spam}}}}, \mu_j^{\textrm{non-spam}})$. Finally, because this distance requires selecting $K$ as an input, we compute distances for each $K$ value in the range $[1, 100]$ to get a holistic view of the ``clusterability" of spam vs. non-spam reports.

The results for each of our distance measures are shown in Figure \ref{fig_distances}. The results confirm our hypothesis that spam reports are less well-structured semantically than non-spam reports. This finding is shown in Figure \ref{fig_distances}b, which shows that spam report embeddings are a considerable distance further from the mean non-spam embedding than are non-spam reports. This indicates that the spam reports are not very well-disguised among the non-spam reports. Moreover, in Figure \ref{fig_distances}c, the distance from spam embeddings to their cluster centers is greater than that of non-spam embeddings for every value of $K$. This suggests our results are robust to number-of-clusters: non-spam reports can be clustered more coherently than spam reports. The results for average within-category distance (Figure \ref{fig_distances}a) are somewhat surprising, however. Though distance within spam reports is greater than distance within non-spam reports (as predicted), the disparity between the two is small. This suggests that despite covering a much wider range of semantic topics than non-spam reports (as outlined in Figures \ref{fig_distances}b and \ref{fig_distances}c), the text within each topic tend to be relatively similar. This is an interesting insight into the behavior of spammers of the VOICE platform; and possibly of spammers in a larger context. The spammers seem to agree upon a set of coordinated spam campaigns that use semantically similar verbiage---a testament of spammers' ability to coordinate attacks.

\section{Conclusion}
This work has initiated a discussion on the security and privacy implications of crowdsourcing and reporting technologies. We have outlined the web of actors, attackers, and victims that are involved in reporting and introduced a unique parallel threat model for this setting. We also examined a case of attacks in the context of a real-world collective spamming attack on the VOICE reporting system. Our analysis suggests that while spam attacks can be harmful to a platform, not all attacks are equal: those that better resemble the set of true reports may be more harmful to an agency's ability to classify true reports from false reports. Humor, troll behavior, and easily-identifiable spam, while more likely to go viral on social media, may be less effective if the goal is obfuscation. As attacker capabilities improve, we posit that deception spam campaigns will pose a significant threat to crowdsourcing technologies. Further research is needed to see how large-scale campaigns, especially those that leverage language models like GPT-3, might be able to submit large numbers of reports that resemble the set of true reports and do not use easily identifiable language.





\bibliography{bib}

\begin{thebibliography}{35}
\providecommand{\natexlab}[1]{#1}
\providecommand{\url}[1]{\texttt{#1}}
\expandafter\ifx\csname urlstyle\endcsname\relax
  \providecommand{\doi}[1]{doi: #1}\else
  \providecommand{\doi}{doi: \begingroup \urlstyle{rm}\Url}\fi

\bibitem[Barocas \& Levy(2020)Barocas and Levy]{barocas2020privacy}
Barocas, S. and Levy, K.
\newblock Privacy dependencies.
\newblock \emph{Wash. L. Rev.}, 95:\penalty0 555, 2020.

\bibitem[Bhatti et~al.(2019)Bhatti, Shah, Maple, and Islam]{bhatti2019novel}
Bhatti, F., Shah, M.~A., Maple, C., and Islam, S.~U.
\newblock A novel internet of things-enabled accident detection and reporting
  system for smart city environments.
\newblock \emph{Sensors}, 19\penalty0 (9):\penalty0 2071, 2019.

\bibitem[Birkbak(2018)]{birkbak2018}
Birkbak, A.
\newblock Into the wild online: Learning from internet trolls, 2018.
\newblock URL
  \url{https://www.firstmonday.org/ojs/index.php/fm/article/view/8297}.

\bibitem[Braithwaite et~al.(2008)Braithwaite, Westbrook, and
  Travaglia]{braithwaite2008attitudes}
Braithwaite, J., Westbrook, M., and Travaglia, J.
\newblock Attitudes toward the large-scale implementation of an incident
  reporting system.
\newblock \emph{International journal for quality in health care}, 20\penalty0
  (3):\penalty0 184--191, 2008.

\bibitem[Braithwaite et~al.(2010)Braithwaite, Westbrook, Travaglia, and
  Hughes]{braithwaite2010cultural}
Braithwaite, J., Westbrook, M.~T., Travaglia, J.~F., and Hughes, C.
\newblock Cultural and associated enablers of, and barriers to, adverse
  incident reporting.
\newblock \emph{BMJ Quality \& Safety}, 19\penalty0 (3):\penalty0 229--233,
  2010.

\bibitem[Brunton \& Nissenbaum(2013)Brunton and
  Nissenbaum]{brunton2013political}
Brunton, F. and Nissenbaum, H.
\newblock Political and ethical perspectives on data obfuscation.
\newblock In \emph{Privacy, due process and the computational turn}, pp.\
  185--209. Routledge, 2013.

\bibitem[Brunton \& Nissenbaum(2015)Brunton and
  Nissenbaum]{brunton2015obfuscation}
Brunton, F. and Nissenbaum, H.
\newblock \emph{Obfuscation: A user's guide for privacy and protest}.
\newblock Mit Press, 2015.

\bibitem[Campello et~al.(2013)Campello, Moulavi, and
  Sander]{campello2013density}
Campello, R.~J., Moulavi, D., and Sander, J.
\newblock Density-based clustering based on hierarchical density estimates.
\newblock In \emph{Pacific-Asia conference on knowledge discovery and data
  mining}, pp.\  160--172. Springer, 2013.

\bibitem[Clark et~al.(2020)Clark, Brudney, Jang, and Davy]{clark2020advanced}
Clark, B.~Y., Brudney, J.~L., Jang, S.-G., and Davy, B.
\newblock Do advanced information technologies produce equitable government
  responses in coproduction: an examination of 311 systems in 15 us cities.
\newblock \emph{The American review of public administration}, 50\penalty0
  (3):\penalty0 315--327, 2020.

\bibitem[Conneau \& Kiela(2018)Conneau and Kiela]{conneau2018senteval}
Conneau, A. and Kiela, D.
\newblock Senteval: An evaluation toolkit for universal sentence
  representations.
\newblock \emph{arXiv preprint arXiv:1803.05449}, 2018.

\bibitem[Fakhraei et~al.(2015)Fakhraei, Foulds, Shashanka, and
  Getoor]{fakhraei2015collective}
Fakhraei, S., Foulds, J., Shashanka, M., and Getoor, L.
\newblock Collective spammer detection in evolving multi-relational social
  networks.
\newblock In \emph{Proceedings of the 21th acm sigkdd international conference
  on knowledge discovery and data mining}, pp.\  1769--1778, 2015.

\bibitem[Freed et~al.(2018)Freed, Palmer, Minchala, Levy, Ristenpart, and
  Dell]{freed2018stalker}
Freed, D., Palmer, J., Minchala, D., Levy, K., Ristenpart, T., and Dell, N.
\newblock “a stalker's paradise” how intimate partner abusers exploit
  technology.
\newblock In \emph{Proceedings of the 2018 CHI conference on human factors in
  computing systems}, pp.\  1--13, 2018.

\bibitem[Friedman \& Albo(2017)Friedman and Albo]{friedman2017punishing}
Friedman, B.~D. and Albo, M.~J.
\newblock Punishing members of disadvantaged minority groups for calling 911.
\newblock \emph{Policing and Race in America: Economic, Political, and Social
  Dynamics. New Brunswick: Lexington Books}, pp.\  141--162, 2017.

\bibitem[Grootendorst(2020)]{grootendorst2020bertopic}
Grootendorst, M.
\newblock Bertopic: Leveraging bert and c-tf-idf to create easily interpretable
  topics., 2020.
\newblock URL \url{https://doi.org/10.5281/zenodo.4381785}.

\bibitem[Hartigan \& Wong(1979)Hartigan and Wong]{hartigan1979algorithm}
Hartigan, J.~A. and Wong, M.~A.
\newblock Algorithm as 136: A k-means clustering algorithm.
\newblock \emph{Journal of the royal statistical society. series c (applied
  statistics)}, 28\penalty0 (1):\penalty0 100--108, 1979.

\bibitem[Kaghazgaran et~al.(2018)Kaghazgaran, Caverlee, and
  Squicciarini]{kaghazgaran2018combating}
Kaghazgaran, P., Caverlee, J., and Squicciarini, A.
\newblock Combating crowdsourced review manipulators: A neighborhood-based
  approach.
\newblock In \emph{Proceedings of the Eleventh ACM International Conference on
  Web Search and Data Mining}, pp.\  306--314, 2018.

\bibitem[Keats~Citron(2018)]{keats2018sexual}
Keats~Citron, D.
\newblock Sexual privacy.
\newblock \emph{Yale LJ}, 128:\penalty0 1870, 2018.

\bibitem[Kessell et~al.(2009)Kessell, Alvidrez, McConnell, and
  Shumway]{kessell2009effect}
Kessell, E.~R., Alvidrez, J., McConnell, W.~A., and Shumway, M.
\newblock Effect of racial and ethnic composition of neighborhoods in san
  francisco on rates of mental health-related 911 calls.
\newblock \emph{Psychiatric Services}, 60\penalty0 (10):\penalty0 1376--1378,
  2009.

\bibitem[Kopan(2017)]{kopan2017voice}
Kopan, T.
\newblock What is voice? trump highlights crimes by undocumented immigrants.
\newblock \emph{CNN}, 2017.
\newblock URL
  \url{https://www.cnn.com/2017/02/28/politics/donald-trump-voice-victim-reporting/index.html}.

\bibitem[Krebs(2019)]{krebs2019swat}
Krebs, B.
\newblock Man behind fatal 'swatting' gets 20 years, 2019.
\newblock URL
  \url{https://krebsonsecurity.com/2019/03/man-behind-fatal-swatting-gets-20-years/}.

\bibitem[McCammon(2021)]{mccammon2021texas}
McCammon, S.
\newblock What the texas abortion ban does — and what it means for other
  states.
\newblock \emph{NPR}, 2021.
\newblock URL
  \url{https://www.npr.org/2021/09/01/1033202132/texas-abortion-ban-what-happens-next}.

\bibitem[McInnes et~al.(2018)McInnes, Healy, and Melville]{mcinnes2018umap}
McInnes, L., Healy, J., and Melville, J.
\newblock Umap: Uniform manifold approximation and projection for dimension
  reduction.
\newblock \emph{arXiv preprint arXiv:1802.03426}, 2018.

\bibitem[Navarro-Carrillo et~al.(2021)Navarro-Carrillo, Torres-Mar{\'\i}n, and
  Carretero-Dios]{navarro2021trolls}
Navarro-Carrillo, G., Torres-Mar{\'\i}n, J., and Carretero-Dios, H.
\newblock Do trolls just want to have fun? assessing the role of humor-related
  traits in online trolling behavior.
\newblock \emph{Computers in Human Behavior}, 114:\penalty0 106551, 2021.

\bibitem[{\"O}zkul \& {\c{C}}apuni(2018){\"O}zkul and
  {\c{C}}apuni]{ozkul2018police}
{\"O}zkul, M. and {\c{C}}apuni, I.
\newblock Police-less multi-party traffic violation detection and reporting
  system with privacy preservation.
\newblock \emph{IET Intelligent Transport Systems}, 12\penalty0 (5):\penalty0
  351--358, 2018.

\bibitem[Rayana \& Akoglu(2015)Rayana and Akoglu]{rayana2015collective}
Rayana, S. and Akoglu, L.
\newblock Collective opinion spam detection: Bridging review networks and
  metadata.
\newblock In \emph{Proceedings of the 21th acm sigkdd international conference
  on knowledge discovery and data mining}, pp.\  985--994, 2015.

\bibitem[Reimers \& Gurevych(2019)Reimers and Gurevych]{reimers2019sentence}
Reimers, N. and Gurevych, I.
\newblock Sentence-bert: Sentence embeddings using siamese bert-networks.
\newblock \emph{arXiv preprint arXiv:1908.10084}, 2019.

\bibitem[Rock(2019)]{rock2019one}
Rock, J.
\newblock One call away: 911 abuse as a weapon against minorities.
\newblock \emph{FAU undergraduate law journal}, 1:\penalty0 160--160, 2019.

\bibitem[Sasson et~al.(2015)Sasson, Haukoos, Ben-Youssef, Ramirez, Bull, Eigel,
  Magid, and Padilla]{sasson2015barriers}
Sasson, C., Haukoos, J.~S., Ben-Youssef, L., Ramirez, L., Bull, S., Eigel, B.,
  Magid, D.~J., and Padilla, R.
\newblock Barriers to calling 911 and learning and performing cardiopulmonary
  resuscitation for residents of primarily latino, high-risk neighborhoods in
  denver, colorado.
\newblock \emph{Annals of emergency medicine}, 65\penalty0 (5):\penalty0
  545--552, 2015.

\bibitem[Smith \& Holmes(2003)Smith and Holmes]{smith2003community}
Smith, B.~W. and Holmes, M.~D.
\newblock Community accountability, minority threat, and police brutality: An
  examination of civil rights criminal complaints.
\newblock \emph{Criminology}, 41\penalty0 (4):\penalty0 1035--1064, 2003.

\bibitem[Son \& Shmatikov(2010)Son and Shmatikov]{son2010hitchhiker}
Son, S. and Shmatikov, V.
\newblock The hitchhiker’s guide to dns cache poisoning.
\newblock In \emph{International Conference on Security and Privacy in
  Communication Systems}, pp.\  466--483. Springer, 2010.

\bibitem[Sveen et~al.(2007)Sveen, Sarriegi, Rich, and
  Gonzalez]{sveen2007toward}
Sveen, F.~O., Sarriegi, J.~M., Rich, E., and Gonzalez, J.~J.
\newblock Toward viable information security reporting systems.
\newblock \emph{Information Management \& Computer Security}, 2007.

\bibitem[Thomas et~al.(2021)Thomas, Akhawe, Bailey, Boneh, Bursztein, Consolvo,
  Dell, Durumeric, Kelley, Kumar, et~al.]{thomas2021sok}
Thomas, K., Akhawe, D., Bailey, M., Boneh, D., Bursztein, E., Consolvo, S.,
  Dell, N., Durumeric, Z., Kelley, P.~G., Kumar, D., et~al.
\newblock Sok: Hate, harassment, and the changing landscape of online abuse.
\newblock \emph{IEEE Symposium on Security \& Privacy}, 2021.

\bibitem[Vincent et~al.(2021)Vincent, Li, Tilly, Chancellor, and
  Hecht]{vincent2021data}
Vincent, N., Li, H., Tilly, N., Chancellor, S., and Hecht, B.
\newblock Data leverage: A framework for empowering the public in its
  relationship with technology companies.
\newblock In \emph{Proceedings of the 2021 ACM Conference on Fairness,
  Accountability, and Transparency}, pp.\  215--227, 2021.

\bibitem[Wikipedia()]{wikitroll}
Wikipedia.
\newblock Internet troll.
\newblock URL \url{https://en.wikipedia.org/wiki/Internet_troll}.

\bibitem[Zou et~al.(2019)Zou, Xi, Wang, Lu, and Xu]{zou2019reportcoin}
Zou, S., Xi, J., Wang, S., Lu, Y., and Xu, G.
\newblock Reportcoin: A novel blockchain-based incentive anonymous reporting
  system.
\newblock \emph{IEEE Access}, 7:\penalty0 65544--65559, 2019.

\end{thebibliography}
\bibliographystyle{icml2022}

\newpage
\appendix
\onecolumn

\section{Additional Related Work}
Novel crowdsourced reporting systems have been proposed in many domains by computer scientists. Papers have suggested and analyzed healthcare self-reporting technologies \cite{braithwaite2010cultural, braithwaite2008attitudes}, blockchain approaches to anonymous reporting \cite{zou2019reportcoin}, decentralized traffic and accident reporting \cite{ozkul2018police, bhatti2019novel}, and a variety of updates to crime reporting technologies. Studies have attempted to use simulation to see how robust reporting platforms are to scaling reports \cite{sveen2007toward}, however few or no measurement studies have attempted to formally report on the unique security and privacy challenges that these systems face. As crowdsourced sensing and reporting systems become a prominent way for policy-makers to represent constituents and respond to urgent needs, these systems need to be scrutinized from a security perspective to identify the numerous ways that they may enable harms.

Broadly, we leverage the concept of obfuscation conceived in \citet{brunton2013political} and further developed in \citet{brunton2015obfuscation}. The book on the subject discusses a relatively long list of examples of obfuscation in the wild, and does not include reporting mechanisms. We therefore extend the theoretical work on obfuscation to a new domain, and test some empirical questions related to effective and ineffective obfuscation strategies. The reporting technology domain is unique because it involves technology-enabled privacy violations that are ultimately carried out by citizens, rather than the police or state actors. This private arena of privacy violations has been theorized from legal and sociological perspectives, for example in work on privacy contingencies \cite{barocas2020privacy} and intimate partner privacy \cite{keats2018sexual, freed2018stalker}. The moral complexity of this domain--including the questions about harmful use and legitimate misuse of platforms--requires that we extend the taxonomy work in \citet{thomas2021sok} to specifically understand use and abuse behaviors in the context of reporting.

Our taxonomy and threat modelling contributions build on computer security research that systematically analyzes violence including hate and harassment mediated over the internet \cite{thomas2021sok} and other technologies \cite{freed2018stalker}. Our contributions related to the VOICE logs, which aim to characterize broad-participation obfuscation techniques, is most similar to spam detection literature, especially \cite{rayana2015collective, fakhraei2015collective, kaghazgaran2018combating}. These studies tend to focus on rating reviews which include a numerical parameter. They also often explicitly model social networks to understand the spread of information or opinions. We are less interested in how obfuscation campaigns spread, and more interested in the security implications of these campaigns and what makes them detectable and/or powerful.

\section{Clustering Analysis Continued}
\label{apdx_clustering}
In addition to the corpus-wide clustering analysis outlined in Section \ref{sec_clustering}, we perform an additional clustering analysis after manually separating the sets of spam and true reports. 

Formally, given a , we pass each sentence $s_i \in S$ through the embedding-generating function $f_\theta : S \rightarrow \mathbb{R}^d$ (with parameters $\theta$), which produces a corresponding $d$-dimensional embedding vector $v_i \in R^d$. We separate the distribution of report sentences $S$ into distributions for spam ($S_{\textrm{spam}}$) and non-spam ($S_{\textrm{non-spam}}$) and cluster each individually. We again follow the clustering method introduced by \citet{grootendorst2020bertopic}, which yields a set of topics $T_s$ and $T_m$ for spam reports and non-spam reports, respectively.

Our hypothesis is that spam reports will cover a much broader set of conversational topics than non-spam reports -- i.e. $|T_\textrm{spam}| > |T_\textrm{non-spam}|$. This stems from the observation that spam responses may be incited by any of the sub-categories of deceptive and abnormal reporting (as outlined in Sections 3.1 and 3.2), whereas valid reports are constrained to the few topics for which the platform was originally intended. 
\begin{figure*}[t!]
    \centering
    \includegraphics[width=0.95\textwidth]{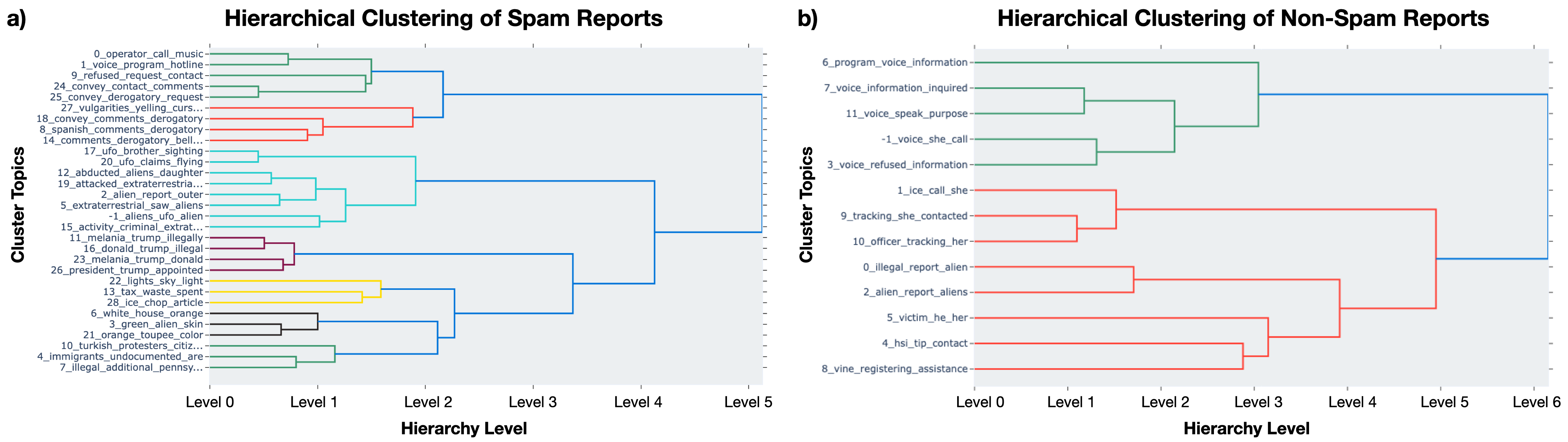}
    \caption{\textbf{a)} Visualization of the $34$ clusters from the set of all spam topics are shown along with their hierarchy. Spam topics include everything from operator harassment (e.g. derogatory comments) to policitally-charged commentary (e.g tax spending, comments about the president) to obviously false reports (e.g. mentions of extraterrestrials). \textbf{b)} Visualization of the $13$ clusters from the set of all non-spam topics. These clusters are more relevant to the intended use of the platform (e.g. information gathering).}
    \label{fig_clustering}
\end{figure*}

Results of this clustering are shown in Figure \ref{fig_clustering}. We find that the optimal number of clusters, as dictated by HDBSCAN, are $|T_\textrm{spam}| = 34$ and $|T_\textrm{non-spam}| = 13$, indicating a much wider range of dialogue in spam reports. Similar to Figure \ref{fig_clustering_all}, we see a large portion of spam reports on the y-axis of Figure \ref{fig_clustering}a dedicated to abnormal reporting---e.g. obviously false reports describing UFO and extraterrestrial encounters (trolling), politically-charged commentary (opinions), and operator harassment (threats, profanity)---and overloading---e.g. staying silent or playing music (report spam). Non-spam reports, on the other hand, consist mostly of information gathering requests and reports of possible tips / witness accounts related to the original use-case of the VOICE system. Altogether, these results confirm our hypothesis from Section \ref{sec_clustering} that spam reports cover a much broader set of conversational topics than non-spam reports.

\section{Structural Analysis of Reports}
\label{apdx_struct}
\begin{figure}[h]
    \centering
    \makebox[\linewidth][c]{\includegraphics[width=0.95\linewidth]{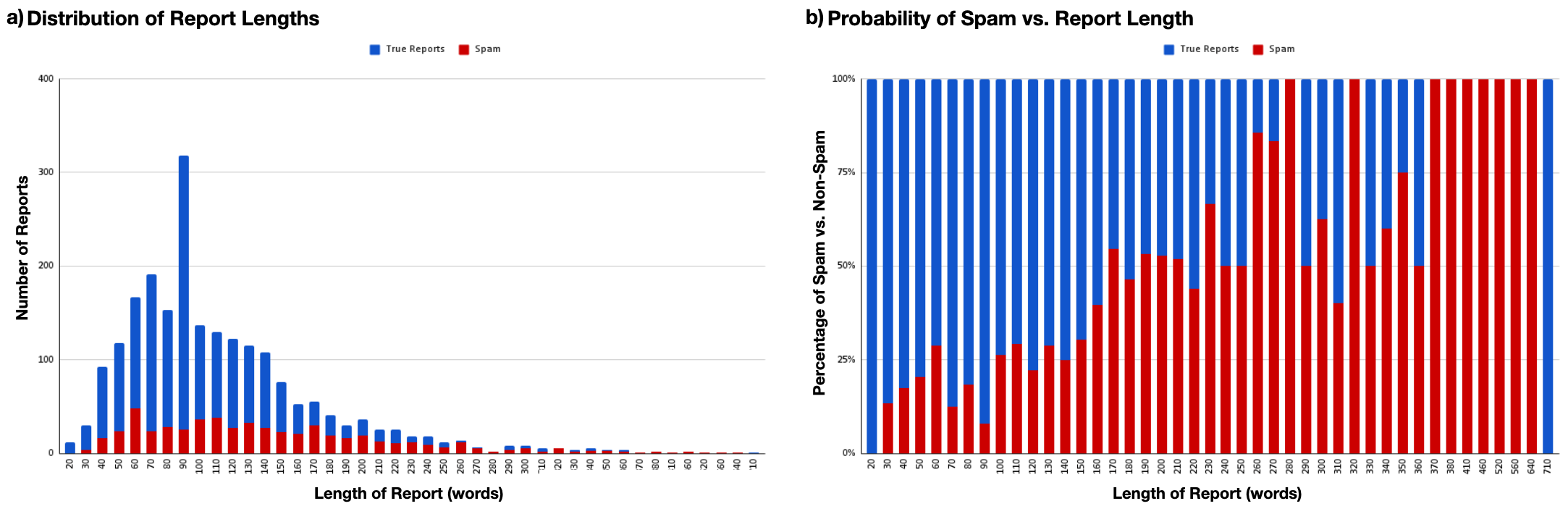}}
    \caption{Analysis of report length for spam (red) vs. non-spam (blue) reports. There exists a positive correlation between report length and the probability that a report is spam.}
    \label{fig_structure_analysis}
\end{figure}

As a first test to compare spam and non-spam activity, we examine the structure (rather than content) of the reports. Given that valid users interact with the platform for legitimate purposes, we expect that non-spam will be more concise and to-the-point than spam. To examine this hypothesis, we measure each report's length in words and compare the relative word counts of spam and non-spam reports.

The results of this study are shown in Figure \ref{fig_structure_analysis}. In Figure \ref{fig_structure_analysis}a, we find that, below $~150$ words, the number of non-spam reports far outweighs the number spam reports. When the number of words in the report exceeds $150$, we find significantly fewer non-spam reports and a larger proportion of spam. This finding is re-iterated in \ref{fig_structure_analysis}b, which shows the probability of a report being spam vs. non-spam, given its length. Again we see a stark shift in the percentage of spam reports when the word count exceed 150 words. Altogether, these results suggest that, though it is certainly possible for a shorter length report to be spam, the likelihood of a report being spam grows considerably with the length of the report.


\end{document}